# Detecting key Soccer match events to create highlights using Computer Vision

Narayana Darapaneni[1], Prashant Kumar[2], Nikhil Malhotra[3], Vigneswaran Sundaramurthy[4], Abhaya Thakur[5], Shivam Chauhan[6], Krishna Chaitanya Thangeda[7], Anwesh Reddy Paduri[8]

[1] Northwestern University/Great Learning, Evanston, US

[2-8] Great Learning, Bangalore, India

anwesh@greatlearning.in

**Abstract.** The research and data science community has been fascinated with the development of automatic systems for the detection of key events in a video. Special attention in this field is given to sports video analytics which could help in identifying key events during a match and help in preparing a strategy for the games going forward. For this paper, we have chosen Football (soccer) as a sport where we would want to create highlights for a given match video, through a computer vision model that aims to identify important events in a Soccer match to create highlights of the match. We built the models based on Faster RCNN and YoloV5 architectures and noticed that for the amount of data we used for training Faster RCNN did better than YoloV5 in detecting the events in the match though it was much slower. Within Faster RCNN using ResNet50 as a base model gave a better class accuracy of 95.5% as compared to 92% with VGG16 as base model completely outperforming YoloV5 for our training dataset. We tested with an original video of size 23 minutes and our model could reduce it to 4:50 minutes of highlights capturing almost all important events in the match.

## 1    Introduction

One of the key requirements in the sporting broadcasting industry is to create highlights of the matches for consumption on streaming platforms or even broadcasting on the channels. Identification of key moments and interesting aspects of the game is currently done mostly manually. This can be time-consuming to detect the sections of the game that can be classified as a highlight and select the relevant frames in a video to show the highlight sequence. This is the problem that we wish to solve through this study. Our idea is to build a model or a combination of models that can help in identifying key moments in a sports match to enable creating highlights automatically and hence much faster than currently being done. The key challenge in sports event detection through computer vision is to identify all possible events that classify as important events worthy of highlights. Each match could be different from the other and the pattern or sequence of movements in a sport like soccer can be highly unpredictable. Having said that, there are specific events that are more important than others. Like a corner kick or penalty kick is more important than a ball pass. Our solution can help video editors spend more time in post-processing and graphics editing of the selected highlight sequence. Reducing the time to production is our key aim which should also help the video editors in spending more time on creative aspects than on non-value-add activities like identifying the events in the match.

## 2 Literature Survey

Our aim is to create highlights of the game effortlessly. There are few companies that offer "AI Cameras " with these capabilities, but the cameras are expensive. A more economical option would be to have your own good resolution cameras and custom model that connects to one or multiple cameras and does processing in the cloud. This could be a more scalable solution that can be extended to sports, outside of soccer as well. For this study, our focus is on soccer. Sports analytics have a lot of applications ranging from broadcasting to coaching and statistical analysis with respect to player's positions during or post live matches. Computer vision plays a key role in these sports analytics, helping coaches in training and providing help to referees during the game (ball tracking, prediction motions, etc.). For broadcasters, tracking and following player's and team's motions, showing the relative position of balls is widely used for TV presenters and commentators.

There is a lot of literature around object detection and action recognition in sports that aid the above-mentioned applications. The majority of the current applications focus on player and ball tracking in real-time to help the broadcasters and TV presenters. But there is little work we came across that is focused on generating highlights of a match. The difference here compared to the other studies is that rather than focusing on a faster real-time object and action detection with lesser accuracy, it is important to get an accurate classification of multiple key actions in a soccer match even if it is not real-time. This will enable the automatic creation of highlights without much manual intervention. Until 2016 majority of the studies focussed on convolutional neural networks for object recognition and action detection. Typically neural networks work in a two-step process. They first try to extract features that are specific to a particular action and then use classification to identify the specific actions.

Studies like [1] have evaluated various convolutional neural network models for action detection in football. Based on their study, a new pooling method (temporal pyramid pooling) to consider the temporal component in sports actions was one of the best models to detect sports events. They split the video into different segments and frames are sampled from the segments. Instead of extracting features from the frames directly, they used a pyramid pooling layer on top which helped the network incorporate temporal granularity scopes. They called this method Deep networks with Temporal Pyramid Pooling (DTPP).

Region-based convolutional neural networks R-CNN brought significant advances in object detection. But they were computationally very expensive. The Fast R-CNN model [11] achieved near real-time results using very deep networks and sharing convolutions across regional proposals. More evolved version, the Faster R-CNN model [13] uses Regional Proposal Networks (RPNs) which leverages the attention mechanism of neural networks and proposes a region. A second unified module uses a Faster R-CNN detector that works on the proposed region. These two networks share the convolution layers which improves the performance of the model.

After 2016 Yolo, Yolov3, Yolov4 family of object detection models [9], [14], [17], [18] were introduced and became popular as they are more performant with real-time predictions. There is literature comparing R-CNN and Yolo models [11] which concludes Yolov3 provides a good tradeoff between speed and accuracy. The Yolo (You Only Look Once) family uses a single convolution network that predicts multiple bounding boxes and classes of those boxes. Unlike R-CNN it doesn't use regions of the image. Rather it uses features from the entire image to predict each bounding box and the associated classes.

For the purpose of this study, we will evaluate the popular R-CNN and Yolo models to see which model works well for our specific application which is to create highlights of a match video. Our unique focus area is to create a model that can identify multiple important actions and events of a football match, rather than focusing on a ball or player tracking.

## 3    Methodology and approach

We created a training set of images consisting of each of the key events. We downloaded the following sample of images and manually annotated them using labelImg:

1.  class foul: 27 images
2.  class corner kick: 89 images
3.  class goal: 57 images
4.  class penalty kick: 81 images

We applied horizontal flip as the only data augmentation technique for this project.

We used object detection methodology to identify the important events in a football match. Primarily we used Faster RCNN with VGG16 as our base model to train using the above-mentioned dataset of 231 images.

We later evaluated performance against a Faster RCNN model built from ResNet50 as the base model and also against YoloV5. The details of which are captured in later sections. Ultimately using the trained model we ran a test video of 23 mins to successfully reduce it to 4:50 mins as highlights.

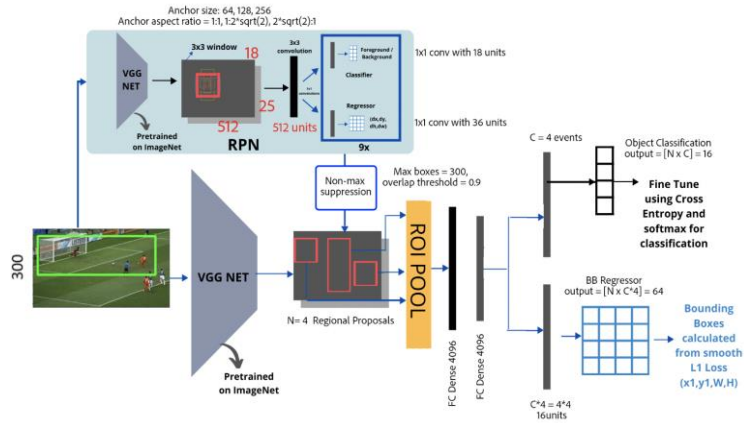

**Fig. 1.** Illustrative picture of the architecture of the Faster RCNN model

### 3.1 Dataset used for the Analysis

A key part of the highlights of any soccer match is identifying important moments that need to be captured. We examined all highlight videos that are currently available. Post this exercise, for this study, we listed out what the significant moments were - 1. Red and Yellow cards for fouls, 2. Goal kicks/attempt at goal, 3. Corner Kicks, and 4. Penalty kicks.

We converted the videos into image frames. For training, we filtered images/frames that we collected for the events specified. In order to effectively annotate them, we used labelImg, a graphical image annotation tool used for label object bounding boxes in images. The image classification and bounding boxes would be input data to train the models. We created a training set of images consisting of each of the key events. Our final model used a dataset of 254 images consisting of the four events mentioned above.

### 3.2 Exploratory Data Analysis / Data Pre-processing

The Images collected were of different sizes. We resized them to a min dimension of 300 px (before feeding into the training model). We also did visualization with bounding boxes. We have used data augmentation techniques to artificially expand the size of a training dataset by creating modified versions of images in the dataset. We have used the image flip technique for data augmentation in this project. For photographs like the football match photograph used in this project, horizontal flips would make sense hence we have used the horizontal flip technique. We generated train.csv and test.csv by combining images and annotation files. Then converted the data frames into annotation.txt files which would be used in model training later.

We used RoboFlow https://app.roboflow.com/ website for uploading images and annotating as required by RoboFlow format to be used in sample YoloV5 model shared on the roboflow website.

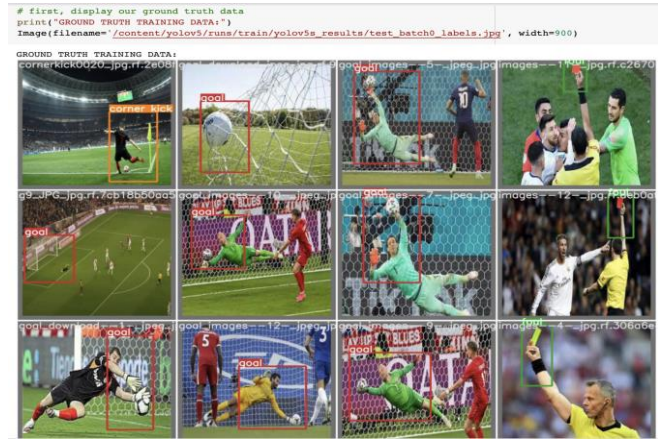

**Fig. 2.** Samples of training images and their annotations

### 3.3 Models and Architecture used

We restricted our project to the most accurate algorithms based on insights from literature survey as our project is not about real time prediction rather identifying an event more accurately for post live highlights creation. Based on the analysis of performance of the state of the art algorithms for object detection, Faster RCNN and Yolo v3 provide the best accuracy [11]. Some recent models have better performance in terms of speed of model evaluation but they slightly compromise on the accuracy. For the purpose of our project we restricted our evaluation to the following 3 models in the order of our project execution.

1. Faster RCNN (VGG as base model) [13]
   We used pre-trained weights for the VGG-16 model downloaded from this GitHub resource. The input image's smallest size is considered as 300 pixels in our case.

2. Yolo v5

3. Faster RCNN (ResNet50 as base model)
   The details and parameters we used in our model for Faster RCNN are exactly the same as what we used with ResNet50 model except for the differences called out below:

   - Input size of the image was reshaped to 320x320 for alignment to use trained weights of ResNet50
   - The output features generated by the base model ResNet50 in this case was 2048 which is the same as the output features from stage 5 block.

### 3.4 Model building and process

We took an iterative approach to improve the model over multiple attempts. Initially we started off with a simple model of Faster RCNN with VGG16 as the base model and with only 2 events - corner kicks (45 images) and fouls (27 images) to evaluate how the model would train and identify test images.

We noticed that the model is not accurate in predicting the important events. Based on our observations, we decided to increase the bounding box threshold to 0.9 to not qualify the false positives bounding boxes and also improve annotations and add more events data for better training.

Then using the same model we increased our training images to 131 and included all the four events. At this point the event detection accuracy seemed to increase. The only penalty kick in the match was correctly identified as shown in the fig.3 below. However some images, especially goals were way off in terms of prediction as shown in fig.4. Also, out of the 3 goals that were scored in the match only 1 was captured in video highlight.

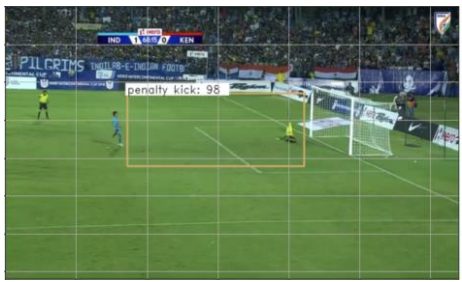 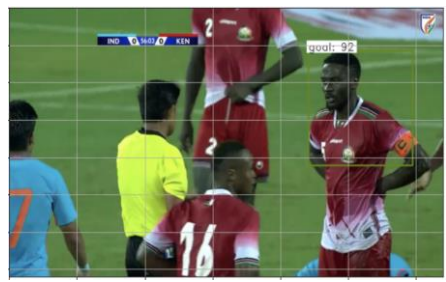

**Fig. 3.** Identification of penalty kick      **Fig. 4.** Player marked as a goal

We realized the images of goals required re-annotation and also increased the number of images. The same model (FRCNN with VGG as base model) was re-ran on 243 images now. This time the predictions were more streamlined and we were almost able to capture all important events in the match. Some false positives were still there in the output as shown below in Fig. 5. However, there were also correct predictions as shown in Fig.6:

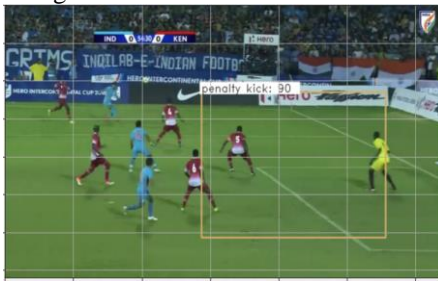 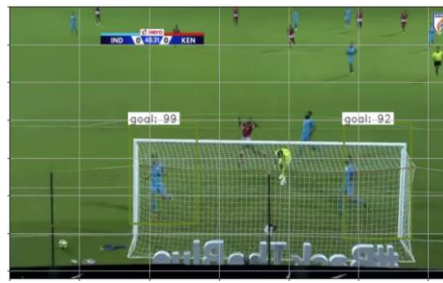

**Fig. 5.** Identified incorrectly as a penalty kick      **Fig. 6.** Identified correctly as goal

Now we changed the base model to see if sophistication can improve model performance. We used ResNet50 as our base model and re-ran the training with the same dataset (243 training images). The training performance was very encouraging as we got 95% accuracy. When we tested on our test video there were hardly any false positives at 0.9 bounding box confidence level.

This model correctly predicted a lot of the important events as shown below in Fig.7. Corner kicks still had false positives as shown in Fig.8:

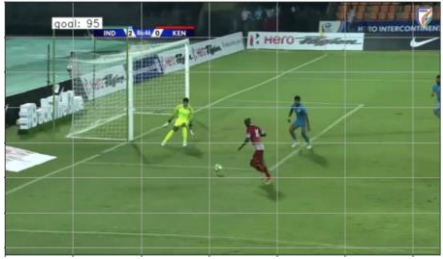  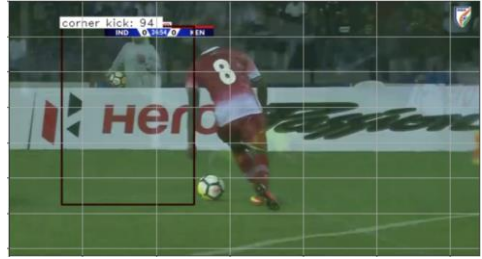

**Fig. 7.** Identified as goal    **Fig. 8.** Identified as a corner kick

We then wanted to compare the performance with more state-of-the-art object detection algorithms in real-time the Yolo series. We used a publicly available jupyter notebook from roboflow to try out YoloV5. The training images used were the same that we used in the earlier 4th and 3rd passes. The results unfortunately weren't promising for YoloV5. The precision and accuracy for each class was very low and it looked like the model required far more images to train effectively. The model took very less time to run but overall accuracy was extremely low. Out of 21 test images the model could identify only 1 image (foul).

Once we got initial results from the models we tried out different hyperparameters like changing the threshold of the bounding box probability and bounding box overlap threshold to verify the results. From our observations from multiple runs on the test video we observed that the following parameters worked best for the respective models:

For Faster RCNN with VGG16

- 0.9 box classification confidence value. (A lot of false positives when we tried to decrease the confidence value)
- 0.7 RPN to ROI overlap threshold.

For Faster RCNN with ResNet50

- 0.6 box classification confidence value. With higher confidence values true events were getting missed.

### 3.5 Test video and generating output highlights

Our objective was to build a highlight video from a real football match after the training model was built. We used the pickled model weights after training to run on test video to generate the event prediction with probability and use the predicted frames to generate highlights. The detailed approach we took is mentioned below.

In order to simplify the process we have created a processing pipeline which makes 3 passes on the video to provide the desired output.

<u>First pass - Preprocess video</u>: In this pass we have used OpenCV and extracted frames as images every 2 seconds. While writing the image to the disk we also passed the time information in the file name with an underscore separator.

<u>Second pass - Predict the event</u>: In this pass we picked the images generated from the previous phase in a list. We ensured that these images are sorted chronologically based on the time information present in the file name. Then we re-formatted the images like resizing based on the model's desired input and called the classifier model's predict() function. Then we used the output of the predict to map to correct classification and confidence. Using the confidence we only filtered the events which have values higher than 90%. By the end of this pass we create a metadata file(tab separated) of the video we have processed which contains details like at what second an event has occurred and what is the confidence of those events. Below is the screenshot of the file. In the below file the first column is the seconds at which an event has happened and the second column contains the class of the event and the confidence score.

```
86 -[('foul', 92.54742860794067)]
88 -[('foul', 90.17170500755311)]
96 -[('Corner kick', 91.70153737068176)]
98 -[('Corner kick', 94.42753791809082)]
112 -[('foul', 97.26079106330072)]
114 -[('Corner kick', 92.61274933915002), ('Corner kick', 91.55545830726624)]
116 -[('foul', 91.55250191685538)]
174 -[('Corner kick', 97.12415933609009)]
198 -[('Corner kick', 92.52678751945496)]
222 -[('foul', 93.76696340190308)]
238 -[('Corner kick', 94.10116460655212)]
310 -[('foul', 94.79607343673706)]
312 -[('foul', 94.05723214149475)]
336 -[('Corner kick', 92.01372061862183)]
344 -[('foul', 90.34197330474854)]
346 -[('foul', 93.66563558578491)]
356 -[('Corner kick', 92.30276481628410)]
372 -[('Corner kick', 91.83760033981323)]
300 -[('Corner kick', 91.04745380031006)]
412 -[('foul', 95.09040117263794)]
438 -[('Corner kick', 90.63490033149719)]
442 -[('Corner kick', 90.57803153991699)]
```

**Fig. 9.** Snapshot of metadata file containing event occurrence and their confidence scores

<u>Third pass - Extract the video:</u> Now that we have the metadata file containing the event time information, we processed the file to create the start and end time of the clip for each event. The start time is calculated by adding a few seconds just before the event has happened and the end time is calculated by adding a few seconds just after the event has happened. In cases when there are events that happened within the specified threshold time of the previous event, then a condition has been added to optimize the end time so that both these events are captured within the same clip.

Apart from extracting the video with the start and end time, we extract the class info with the confidence score from the metadata file and then write the class with max confidence score on all the frames of the video. The text is written on the top left corner of the frames as shown in the picture below. We use this in the final video to show what the model predicted that particular frame as. In cases where multiple events were classified in the same frame we picked the one with highest confidence to show in the frame title with class and confidence value.

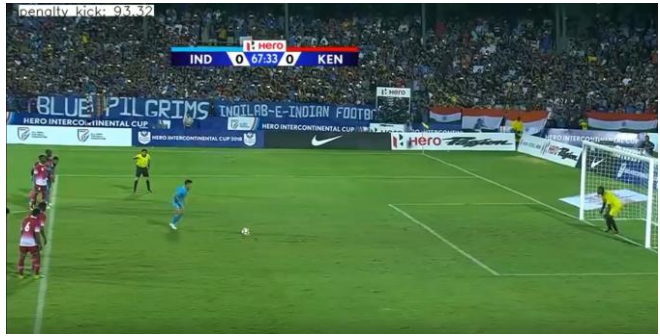

**Fig. 10.** Screenshot of the video with the event class and confidence score on top left corner

At the end of the execution, we generated a video file containing the clips of all the events marked in the metadata file and the class label of the event with confidence score.

## 4 Model evaluation

### 4.1 Results from Faster RCNN (VGG16 base):

We received 92% class accuracy on the training data set with 40 epoch runs consisting of 243 training images. Below are the model parameter progressions as it continues to learn over epochs. Increasing mean_overlapping_bboxes indicate that as the model learns there are more boxes converging towards a ground truth box.

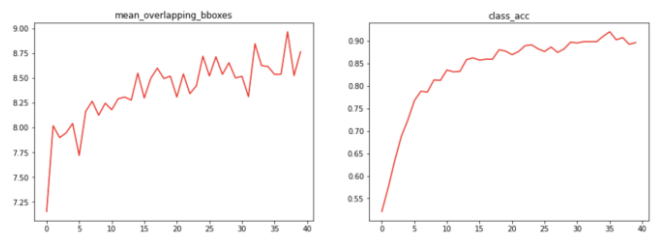

**Fig. 11.** Class accuracy and mean overlapping boxes from Faster RCNN (VGG16 model)

## 4.2 Results from Faster RCNN (ResNet50 base):

We obtained 95.5% accuracy with the ResNet50 base model with 40 epochs on the same 243 training images. You can notice that this model took less time to learn. It reached 85% within 7 epochs as compared to 14 epochs required by the VGG16 base model.

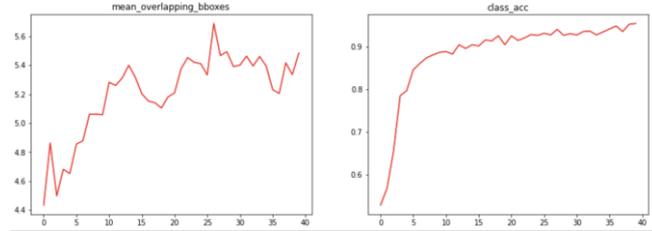

**Fig. 12.** Class accuracy and mean overlapping boxes from Faster RCNN (ResNet50 model)

## 4.3 Results from YoloV5:

Overall precision of the model is very low at 0.238 and recall of 0.406.

**Table 1.**. Performance metrics from YoloV5 model output

```
           Class    Images   Targets        P        R    mAP@.5  mAP@.5:.95: 100% 2/2 [00:01<00:00,  1.55it/s]
             all       45        45    0.238    0.406     0.287     0.113
     corner kick       45         7    0.141    0.286    0.0645    0.0184
            foul       45         9    0.386    0.667     0.513     0.194
            goal       45        14   0.0533   0.0714    0.0274   0.00848
     penalty kick      45        15    0.372      0.6     0.543     0.233
Optimizer stripped from runs/train/yolov5e_results/weights/last.pt, 14.8MB
Optimizer stripped from runs/train/yolov5e_results/weights/best.pt, 14.8MB
100 epochs completed in 0.377 hours.

CPU times: user 17.8 s, sys: 2.59 s, total: 20.4 s
Wall time: 23min 2s
```

**Table 2.** Comparison of training between models

| Model | Class Accuracy | Mean Elapsed time per epoch | Mean elapsed time to evaluate test image |
|---|---|---|---|
| Faster RCNN (VGG base) | 92% | 2.17 mins | 4.8s |
| Faster RCNN (ResNet50 base) | 95.% | 3.4 mins | 5.5s |
| YoloV5 | 30% | 13.58s | 0.653s |

Note: all results mentioned above are evaluated under the GPU environment in Colab.

## 4.4 Performance on the test video

The video contained the following events with corresponding frequencies during the match: Penalty kick – 1; Goals – 3, Corner kick – NIL, Foul - NIL

There were about 7 shots at goal which didn't end up converting as goals but we would expect our model to predict them as goals. So we will include this in our comparison chart.

Following are the correct predictions summary from each of the models we ran. We used 0.9 bounding box classification threshold for Faster RCNN (with VGG) and 0.8 threshold for Faster RCNN (with ResNet50):

**Table 3.** Comparison of performance of models on test video

| Event Name | Actual instances | Instances predicted by Faster RCNN (VGG) | Instances predicted by Faster RCNN (ResNet50) | Instances predicted by YoloV5 |
|---|---|---|---|---|
| Penalty kick | 1 | 1 real + 3 false | 0 | 0 |
| Goals | 3 | 3 real | 2 real | 0 |
| Shots at goal | 7 | 7 real + 10 false | 7 real + 7 false | 0 |
| Corner kick | 0 | 11 false | 11 false | 0 |
| Foul | 0 | 0 | 1 false | 0 |

Best video output size for all the models used:

1. Faster RCNN with VGG16 as base: 4:50 mins
2. Faster RCNN with ResNet50 as base: 30 secs
3. YoloV5: 0 secs

Though Faster RCNN with ResNet50 produced the least false positives it also missed in terms of picking the right event. We believe this could be due to an over-trained model. By providing more diverse images of each event type we believe ResNet50 significantly can outperform the model with VGG16.

Example images of the correct predictions by our model (Faster RCNN with VGG16):

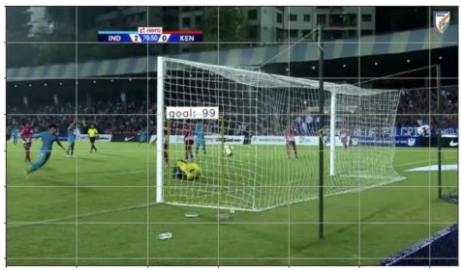 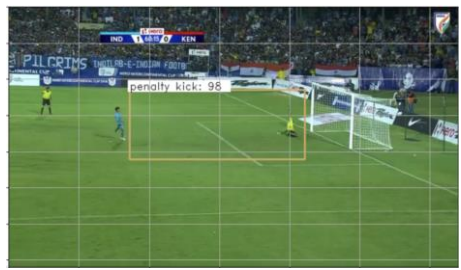

**Fig. 13.** Model capturing 2$^{nd}$ goal    **Fig. 14.** Identified as penalty kick

Ultimately, our best highlights video was of 4:50 min by capturing most important events successfully.

## 5     Limitations and next steps

Major limitation of the project is that we used the approach of using object detection and not sequence of images to identify the actual event in a football match. To enhance the highlight video output we can try to use the LSTM model to identify the sequence and predict actual events better especially like shots at goal or actual goals.

As mentioned above there are instances where an event is classified incorrectly. Below are some reasons that can be attributed to incorrect classification and areas of improvement for our model.

- One of the primary reasons is less training data. Generating the training data for the use case is a very time consuming and manual process. For academic purposes we are currently using less than 80 images per event possibly not covering all camera angles.
- There is a need to add more classes of events in our training set which in turn will categorize the events correctly and also improve our models accuracy.
- Image augmentation can help in increasing the training data. We can consider adding grayscale augmentation for our original training dataset.
- Optimizing the hyperparameters in the model can help improve the  model performance. We can consider changing the number of regions of interest, image size (will increase training time), anchor scales and pool size.

From the perspective of the sports domain, the problem that we were attempting to address was how effectively the key moments in a match can be detected. Sports channels, news and even platforms like YouTube would be benefited to provide a consolidated view of key moments in a game. In terms of the business side of it, the viewer may not be interested in the whole event, however just the moments considered to be the good ones. Through the base model and architecture we have deployed, it allows us to achieve automatic summarization and highlight extraction systems allowing one to watch a few parts of the game focusing on the most interesting moments.